\documentclass[11pt,a4paper]{article}

\usepackage[T1]{fontenc}
\usepackage[utf8]{inputenc}
\usepackage{lmodern}
\usepackage{microtype}

\usepackage[a4paper,margin=1in]{geometry}

\usepackage{graphicx}
\usepackage{booktabs}
\usepackage[numbers,sort&compress]{natbib}
\usepackage[hidelinks]{hyperref}
\usepackage{url}

\usepackage{amsmath,amsfonts,bm}

\def\eqref#1{equation~\ref{#1}}
\def\1{\bm{1}}

\DeclareMathAlphabet{\mathsfit}{\encodingdefault}{\sfdefault}{m}{sl}
\SetMathAlphabet{\mathsfit}{bold}{\encodingdefault}{\sfdefault}{bx}{n}

\title{Variational Neurons in Transformers for Language Modeling}

\author{
Yves Ruffenach \\
Conservatoire National des Arts et M\'etiers \\
Strasbourg, France \\
\texttt{yves@ruffenach.net}
}

\date{}

\begin{document}

\maketitle

\begin{abstract}
Transformers for language modeling usually rely on deterministic internal computation, with uncertainty expressed mainly at the output layer. We introduce variational neurons into Transformer feed-forward computation so that uncertainty becomes part of the internal computation itself. Concretely, we replace deterministic feed-forward units with local variational units based on EVE while preserving the overall Transformer backbone.

We evaluate this design in compact next-token language-modeling settings. We compare deterministic and variational variants with both predictive and probabilistic criteria. Alongside negative log-likelihood, perplexity and accuracy, we analyze calibration, conditional variance, mutual information and latent-usage statistics. The resulting picture is clear. Variational neurons integrate stably into Transformers, preserve strong predictive performance and produce informative uncertainty signals. The experiments also show that task quality, useful depth and internal stability are distinct properties.

These results establish variational Transformers as a practical form of uncertainty-aware language modeling. They show that Transformers can predict with an explicit internal structure of uncertainty, which supports stronger probabilistic evaluation and a more informative analysis of model behavior.
\end{abstract}

\section{Introduction}

Transformer language models are strong predictors, but their internal computations are usually deterministic \citep{vaswani2017attention,radford2019language}. In most current practice, uncertainty is read from the final predictive distribution or recovered through external procedures such as ensembles, dropout, or post-hoc calibration \citep{blundell2015weight,gal2016dropout,lakshminarayanan2017simple,guo2017calibration}. This leaves an important question open. Can uncertainty become part of the computation itself, rather than a quantity observed only at the output?

This question is especially relevant for language modeling. Next-token prediction is inherently multi-valued and many continuations can be plausible. A model that represents uncertainty only at the last layer can still predict well, but it provides limited access to how uncertainty is formed, propagated and used inside the network. We study a complementary design in which uncertainty is introduced directly into the intermediate computation. Our goal is not only to predict what comes next, but to predict with an explicit internal structure of uncertainty.

Our approach builds on EVE, a local variational neuron with a learned posterior, a learned local prior and an explicit control regime for latent activity. We instantiate these neurons inside the feed-forward computation of compact Transformer blocks for next-token language modeling. This preserves the overall Transformer backbone while replacing deterministic intermediate computation with local variational inference. The resulting model combines self-attention with neuron-level stochastic computation, latent-state control and measurable internal uncertainty signals during training.

This design gives us a clean setting to study a practical question. Does local variational computation improve model behavior beyond point prediction? To answer this, we compare deterministic and variational Transformers in compact language modeling experiments with frozen GPT-style embeddings and teacher-forcing windows built from prompt--story pairs. Alongside negative log-likelihood, perplexity and accuracy, we track calibration, mutual information, conditional variance, latent activity and related internal diagnostics. This lets us evaluate not only predictive quality, but also how uncertainty is represented and used during computation.

Our contributions are threefold. First, we introduce a Transformer language model whose feed-forward computation is carried by local variational neurons. Second, we pair this architecture with an explicit latent-control mechanism that keeps the stochastic path active and numerically stable. Third, we provide an uncertainty-centered evaluation protocol for compact next-token language modeling, with metrics that expose both predictive behavior and internal latent usage.

\section{Related Work}

\paragraph{Transformers, uncertainty and latent variables.}
Transformers are the standard architecture for language modeling because self-attention supports strong context modeling and efficient parallel training \citep{vaswani2017attention}. Autoregressive models such as GPT-2 established next-token prediction as a simple and effective training paradigm \citep{radford2019language}. We stay in this setting and change not the task, but the internal feed-forward computation.

A large literature studies uncertainty in neural networks through Bayesian or approximate Bayesian methods, including variational weight inference, Monte Carlo dropout and deep ensembles \citep{blundell2015weight,gal2016dropout,lakshminarayanan2017simple}. These methods improve calibration and predictive confidence estimation, which is important because modern neural networks are often miscalibrated \citep{guo2017calibration}. In parallel, latent-variable models for text have long aimed to capture uncertainty, diversity and higher-level structure, while also highlighting the challenge of posterior collapse \citep{kingma2014auto,bowman2016generating,he2019lagging}. Most of this work uses sequence-level or sentence-level latent variables.

\paragraph{Our position.}
Recent work has also explored uncertainty-aware Transformers through Bayesian dropout variants and Transformer-specific approximate inference schemes \citep{sankararaman2022bayesformer,xiao2020wat}. Our work is complementary but operates at a different level. Rather than attaching uncertainty to weights, recovering it only at the output, or introducing a single global latent code, we place explicit local variational neurons inside Transformer feed-forward blocks. This makes uncertainty part of the forward computation itself and makes latent usage observable throughout the network.

\section{Method}

\subsection{Overview}

We study language modeling with Transformers whose feed-forward computation
uses local variational neurons. Our starting point is EVE, a neuron-level
latent-variable compute primitive. Each neuron maintains a local posterior,
a local prior and an explicit internal control regime. We instantiate this
primitive inside Transformer blocks for next-token prediction. The result is
a language model in which uncertainty is part of the internal computation,
not only a property of the output distribution.

\begin{figure}[t]
    \centering
    \includegraphics[width=1\textwidth]{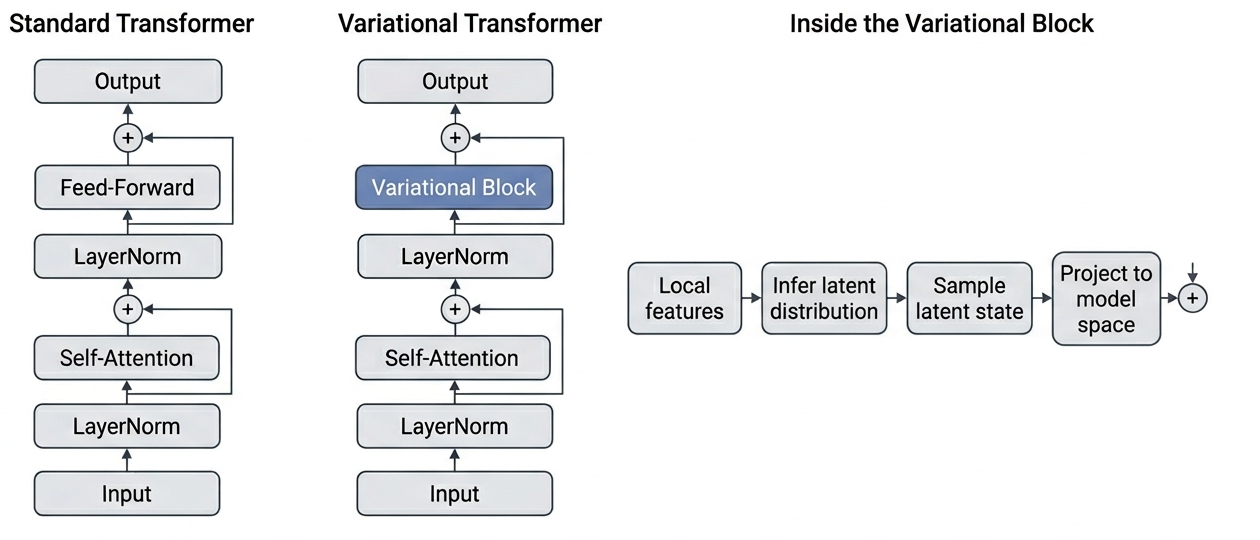}
    \caption{
    Standard and variational Transformer blocks. The overall Transformer backbone is unchanged. Only the feed-forward computation is replaced by a variational block that infers a local latent distribution, samples a latent state and projects it to model space before the residual update.
    }
    \label{fig:variational-transformer-overview}
\end{figure}

Figure~\ref{fig:variational-transformer-overview} summarizes the architectural
change studied in this paper. The Transformer backbone is preserved, while the
deterministic feed-forward computation is replaced by a local variational block.

\subsection{Local variational neuron}

Let \(x\) denote the current neuron input and \(h\) its local memory state.
Following the EVE formulation, each neuron defines a local posterior
\[
q_{\phi}(z \mid x,h)
=
\mathcal{N}\!\big(
\mu_q(x,h),
\operatorname{diag}(\sigma_q^2(x,h))
\big),
\]
and a local prior
\[
p_{\psi}(z \mid h)
=
\mathcal{N}\!\big(
\mu_p(h),
\operatorname{diag}(\sigma_p^2(h))
\big).
\]
The latent state is sampled through the standard reparameterization
\[
z = \mu_q + \sigma_q \odot \epsilon,
\qquad
\epsilon \sim \mathcal{N}(0,I),
\]
and decoded into an activation
\[
y = g_{\theta}(z).
\]

This is the key difference with a deterministic neuron. A deterministic neuron
emits an activation directly from a hidden transformation. Our neuron emits
an activation from a locally inferred latent state. The neuron therefore
carries its own probabilistic internal regime.

\subsection{Local objective, control, and prior}

Each neuron contributes a local variational objective that combines a task
term, a local KL term, and an internal control term:
\[
\mathcal{L}_{\mathrm{local}}
=
\mathcal{L}_{\mathrm{task}}
+
\beta\,
\mathrm{KL}\!\left(
q_{\phi}(z \mid x,h)\,\|\,p_{\psi}(z \mid h)
\right)
+
\mathcal{L}_{\mathrm{control}}.
\]
In our setting, \(\mathcal{L}_{\mathrm{task}}\) is induced by next-token
prediction. The KL term aligns the local posterior with the local prior. The
control term keeps the neuron in a useful operating regime. This objective
supports both learning and interpretation because it regularizes the latent
state while shaping the neuron’s internal behavior.

A central control variable is the latent energy
\[
\mu^2
=
\frac{1}{d_z}
\sum_{j=1}^{d_z} \mu_{q,j}^2,
\]
which measures the average squared posterior mean across latent dimensions.
Very low latent activity weakens the stochastic path, whereas very high
activity destabilizes it. We therefore regulate the neuron around a target
latent-energy regime. At a high level, the neuron is encouraged to remain
near \(\mu^2_{\mathrm{target}}\) and inside an admissible band around that
target. This yields explicit internal statistics such as inside-band mass,
fraction too low, fraction too high, and target gap. In practice, our
Transformer implementation uses homeostatic regulation and band-based
monitoring of latent activity. This keeps the latent path active and makes
its regime directly measurable during training.

The neuron also maintains a local autoregressive prior, which gives temporal
continuity to the latent state and turns the prior into a learned expectation
rather than a fixed isotropic reference. At a high level, the prior mean
evolves as
\[
\mu_p^{(t)}
=
f_{\mathrm{AR}}\!\big(
\mu_p^{(t-1)}, z^{(t-1)}
\big),
\]
with an analogous update for the prior scale when used. This gives the neuron
a short internal memory and lets the current posterior be compared to its own
recent latent trajectory. The result is a local stochastic computation with
explicit latent dynamics rather than a static random activation.

\subsection{Transformer instantiation for language modeling}

We instantiate these local variational neurons inside a compact Transformer
for next-token prediction. Let
\[
X=(x_1,\dots,x_T)
\]
be an input token sequence. Tokens are mapped to embeddings and processed by
a stack of Transformer blocks. Self-attention remains standard. The
feed-forward computation is replaced by local variational neurons.

For block \(\ell\), let \(H^{(\ell)}\) denote the token representations at the
input of the block. We first apply self-attention:
\[
\widetilde{H}^{(\ell)}
=
H^{(\ell)}
+
\mathrm{MHA}\!\left(
\mathrm{LN}(H^{(\ell)})
\right).
\]
We then form a block-level hidden representation \(u^{(\ell)}\) and pass it
through a bank of local variational neurons. For neuron \(i\),
\[
q_{\phi}^{(\ell,i)}(z_i^{(\ell)} \mid u^{(\ell)}, h_i^{(\ell)})
=
\mathcal{N}\!\big(
\mu_{q,i}^{(\ell)},
\operatorname{diag}((\sigma_{q,i}^{(\ell)})^2)
\big),
\]
\[
p_{\psi}^{(\ell,i)}(z_i^{(\ell)} \mid h_i^{(\ell)})
=
\mathcal{N}\!\big(
\mu_{p,i}^{(\ell)},
\operatorname{diag}((\sigma_{p,i}^{(\ell)})^2)
\big),
\]
and
\[
z_i^{(\ell)}
=
\mu_{q,i}^{(\ell)}
+
\sigma_{q,i}^{(\ell)} \odot \epsilon_i,
\qquad
\epsilon_i \sim \mathcal{N}(0,I).
\]

The sampled latent activations are then decoded and projected back to the
model space to produce the feed-forward update. This preserves the overall
Transformer backbone while making the intermediate computation locally
variational.

\subsection{Prediction head}

The final hidden representation is mapped to next-token logits through a
standard language-model head. In the variational model, prediction is
evaluated through Monte Carlo averaging over stochastic forward passes. Let
\[
\{\mathbf{y}^{(m)}\}_{m=1}^{M}
\]
denote the logits obtained from \(M\) stochastic samples and let
\[
\mathbf{p}^{(m)} = \mathrm{softmax}(\mathbf{y}^{(m)}).
\]
The Monte Carlo predictive distribution is
\[
\bar{\mathbf{p}}
=
\frac{1}{M}\sum_{m=1}^{M}\mathbf{p}^{(m)}.
\]
All probabilistic evaluation metrics, including Monte Carlo predictive NLL,
are computed from this predictive distribution. This Monte Carlo view lets us
evaluate both predictive quality and uncertainty-aware behavior.

\subsection{Global training objective}

At the model level, training combines the language-modeling objective with
the local variational and control terms accumulated across units and layers:
\[
\mathcal{L}
=
\mathcal{L}_{\mathrm{LM}}
+
\beta\,\mathcal{L}_{\mathrm{KL}}
+
\mathcal{L}_{\mathrm{control}}
+
\alpha_{\mathrm{AR}}\,\mathcal{L}_{\mathrm{AR}},
\]
where \(\mathcal{L}_{\mathrm{LM}}\) is the next-token prediction loss,
\(\mathcal{L}_{\mathrm{KL}}\) aggregates the local KL terms and
\(\mathcal{L}_{\mathrm{AR}}\) is present when the autoregressive prior is
active. This objective keeps the language-modeling task central while
preserving explicit control over latent activity and latent dynamics.

\subsection{Deterministic baseline and evaluation}

To test whether local variational computation is genuinely useful, we compare
the variational Transformer to a deterministic baseline built from the same
overall architecture. The comparison is therefore not between unrelated
models. It isolates the effect of replacing deterministic feed-forward
neurons with local variational neurons.

We evaluate both models on standard predictive metrics, including negative
log-likelihood, perplexity and accuracy. We complement these with
uncertainty-centered metrics such as calibration, conditional variance,
mutual information and internal latent diagnostics. This matches the goal
of the paper. We do not only ask whether the model predicts well. We ask
whether it predicts with a useful internal structure of uncertainty.

\section{Experiments}

\subsection{Experimental scope and protocols}

We report four controlled experimental blocks on \textit{WritingPrompts-Filtered}. The first block is the 19.9k-example setting, with about 17.9k training examples and 2.0k validation examples, used for Run A and Run B and for the task-quality versus useful-depth comparison. The second block is the 10.0k-example setting, with about 8.0k training examples and 2.0k validation examples, used for the v23 versus v22 stabilization comparison. The third block is a dedicated matched deterministic-versus-variational comparison on 2.0k examples. The fourth block is a separate matched EVE-versus-DET comparison on an approximately 19,925-raw-example protocol with \texttt{val\_frac=0.2}.

Across all blocks, self-attention and the language-model head are kept fixed. The comparison isolates the effect of replacing deterministic feed-forward computation with local variational computation and, where stated, stabilization patches.

\subsection{Evaluation logic}

We evaluate the runs along four complementary axes: language-modeling quality, probabilistic quality, effective depth usage and internal stability. We distinguish three metric sources throughout the section. \textbf{Final validation} reports the optimization objective (\texttt{loss}) together with standard task metrics (\texttt{ce}, \texttt{ppl}, \texttt{acc}). Because \texttt{loss} may include regularization and control terms, task-level comparisons rely primarily on CE, perplexity and accuracy. \textbf{Extended validation} reports uncertainty-aware probabilistic metrics, including Monte Carlo predictive NLL (\texttt{nll}), \texttt{ece}, \texttt{mutual\_information}, \texttt{conditional\_variance\_mc}, \texttt{top1\_flip\_rate\_mc} and \texttt{cvar\_nll}. \textbf{Internal diagnostics} report layer weights and latent statistics, including per-layer KL and latent energy $\mu^2$.

\subsection{19.9k-example setting: task quality and useful depth}

Run A on the 19.9k-example setting is a strong run. In \textbf{final validation}, it reaches
\texttt{loss} $=4.8118$, CE $=4.7370$, perplexity $=114.10$ and accuracy
$=0.2293$. For task-level comparisons, we primarily interpret CE,
perplexity and accuracy.
In \textbf{extended validation}, the run remains genuinely probabilistic, with Monte Carlo predictive NLL $=4.8164$, ECE $=0.0445$, mutual information $=0.1188$, conditional variance $=0.00209$, top-1 flip rate $=0.2757$ and CVaR-NLL $=11.72$. In \textbf{internal diagnostics}, the head is almost entirely dominated by the first layer, with learned layer weights $0.9120/0.0438/0.0442$. The latent statistics reveal a more subtle picture: layer 2 is still genuinely active in latent terms, with $\mathrm{KL}_2=0.1920$ and $\mu^2_2=0.0911$, while layer 3 is essentially dead, with $\mathrm{KL}_3 \approx 1.6\times 10^{-5}$ and $\mu^2_3 \approx 1.5\times 10^{-5}$.
Run A therefore activates a meaningful second latent layer, although most final-head usage remains concentrated in layer 1.

Run B on the 19.9k-example setting slightly improves the task-level metrics. In \textbf{final
validation}, it reaches \texttt{loss} $=4.7950$, CE $=4.7217$, perplexity
$=112.36$ and accuracy $=0.2307$.
In \textbf{extended validation}, it remains probabilistic, with Monte Carlo predictive NLL $=4.8286$, ECE $=0.0457$, mutual information $=0.1168$, conditional variance $=0.00173$, top-1 flip rate $=0.2549$ and CVaR-NLL $=11.82$. The \textbf{internal diagnostics}, however, are much more concentrated than in Run A. The head becomes even more layer-1 dominated, with weights $0.9171/0.0462/0.0367$ and the deeper layers become nearly inactive: $\mathrm{KL}_2=2.0\times 10^{-5}$, $\mu^2_2=1.94\times 10^{-5}$, $\mathrm{KL}_3=1.47\times 10^{-5}$ and $\mu^2_3=1.33\times 10^{-5}$. The active latent fractions are also extremely small, with only $0.0059$ active fraction in layer 2 and $0.0010$ in layer 3.
Run B therefore highlights a clear task--depth trade-off: it brings a small task-level gain while sharply reducing useful deep latent computation.

\begin{table*}[t]
\centering
\caption{19.9k-example setting. Final-validation objective and task metrics, together with extended-validation probabilistic metrics. Run B is slightly stronger on task-level metrics, while Run A preserves the stronger deep latent regime.}
\small
\setlength{\tabcolsep}{4pt}
\begin{tabular}{lcccccccc}
\toprule
Run & Loss$\downarrow$ & CE$\downarrow$ & PPL$\downarrow$ & Acc$\uparrow$ & MC NLL$\downarrow$ & ECE$\downarrow$ & MI & Flip \\
\midrule
Run A (19.9k) & 4.8118 & 4.7370 & 114.10 & 0.2293 & 4.8164 & 0.0445 & 0.1188 & 0.2757 \\
Run B (19.9k) & \textbf{4.7950} & \textbf{4.7217} & \textbf{112.36} & \textbf{0.2307} & 4.8286 & 0.0457 & 0.1168 & 0.2549 \\
\bottomrule
\end{tabular}
\end{table*}

\begin{table*}[t]
\centering
\caption{19.9k-example setting internal diagnostics. Run A keeps a genuinely active second latent layer, whereas Run B almost collapses the deeper layers.}
\small
\setlength{\tabcolsep}{4pt}
\begin{tabular}{lccccccccc}
\toprule
Run & $w_1$ & $w_2$ & $w_3$ & $\mathrm{KL}_1$ & $\mathrm{KL}_2$ & $\mathrm{KL}_3$ & $\mu_1^2$ & $\mu_2^2$ & $\mu_3^2$ \\
\midrule
A (19.9k) & 0.9120 & 0.0438 & 0.0442 & 0.1608 & 0.1920 & $\sim 1.6{\times}10^{-5}$ & 0.1551 & 0.0911 & $\sim 1.5{\times}10^{-5}$ \\
B (19.9k) & 0.9171 & 0.0462 & 0.0367 & 0.1636 & $2.0{\times}10^{-5}$ & $1.47{\times}10^{-5}$ & 0.1522 & $1.94{\times}10^{-5}$ & $1.33{\times}10^{-5}$ \\
\bottomrule
\end{tabular}
\end{table*}

\subsection{10.0k-example setting: raw LM quality and stabilization}

The 10.0k-example block serves a different purpose. Here, the key comparison is between the raw two-phase run v23 and the stabilized layer3aux run v22. In \textbf{final validation}, v23 remains slightly ahead on raw language-modeling metrics, with CE $=4.8157$, perplexity $=123.44$ and accuracy $=0.2256$. In \textbf{internal diagnostics}, however, its deep regime is much less controlled: $\mu^2_{\text{eval}}=1.5313$, $\mu^2_{\text{std,eval}}=53.45$ and the third layer reaches $\mu^2_3=550.82$.
Thus, v23 provides the strongest raw CE/PPL point in this setting, while its internal latent regime remains substantially less controlled.

By contrast, v22 gives up only a negligible amount of raw LM quality. In \textbf{final validation}, it reaches CE $=4.8208$, perplexity $=124.06$ and accuracy $=0.2264$. In \textbf{extended validation}, its probabilistic behavior remains informative, with ECE $=0.0535$, mutual information $=0.1687$, top-1 flip rate $=0.2852$, $\sigma_{\text{mean}}=0.8866$, active unit fraction $=1.0$ and effective active dimensions $=1024$. In \textbf{internal diagnostics}, the head is still strongly layer-1 dominated, with weights $0.9128/0.0425/0.0447$, so the result is not one of balanced depth usage. But the stabilization effect is substantial: $\mu^2_{\text{eval}}$ drops to $0.2119$, $\mu^2_{\text{std,eval}}$ drops to $2.94$ and the third-layer latent energy drops from $550.82$ in v23 to $13.78$ in v22.
The residual excess shifts mostly to layer 2, where $\mu^2_2=37.43$, while the most severe third-layer instability is strongly reduced.

Overall, v23 provides the best raw LM metrics in this setting, whereas v22 provides the cleanest stabilized variational regime.

\begin{table*}[t]
\centering
\caption{10.0k-example setting. Final-validation task metrics and internal diagnostics. v23 is slightly better on raw LM metrics, while v22 is substantially cleaner internally.}
\setlength{\tabcolsep}{4pt}
\begin{tabular}{lcccccc}
\toprule
Run & CE$\downarrow$ & PPL$\downarrow$ & Acc$\uparrow$ & $\mu^2_{\mathrm{eval}}\downarrow$ & $\mu^2_{\mathrm{std}}\downarrow$ & $\mu^2_3\downarrow$ \\
\midrule
v23 (10.0k) & \textbf{4.8157} & \textbf{123.44} & 0.2256 & 1.5313 & 53.45 & 550.82 \\
v22 (10.0k) & 4.8208 & 124.06 & \textbf{0.2264} & \textbf{0.2119} & \textbf{2.94} & \textbf{13.78} \\
\bottomrule
\end{tabular}
\end{table*}

\subsection{Matched deterministic-versus-variational comparison on approximately 19,925 raw examples}

A lower-budget matched comparison already points in the same direction. On the dedicated 2.0k-example setting, both \texttt{run\_variational} and \texttt{run\_deterministic} are enabled, the variational model uses the layer-3 auxiliary patch, and checkpoint selection uses \texttt{select\_by="ce"}. In \textbf{final validation}, EVE reaches CE $=5.2896$, perplexity $=198.26$, and accuracy $=0.1752$, whereas DET reaches CE $=5.5417$, perplexity $=255.12$, and accuracy $=0.1525$. EVE also shows cleaner early learning dynamics, with validation CE improving from $5.4466$ to $5.3315$ to $5.2859$ over the first three epochs, whereas DET peaks earlier. In \textbf{extended validation}, EVE reaches Monte Carlo predictive NLL $=5.4709$, ECE $=0.0444$, mutual information $=0.1830$, conditional variance $=0.00144$, top-1 flip rate $=0.3288$, and CVaR-NLL $=12.51$. DET reaches NLL $=5.6673$, ECE $=0.0603$, and CVaR-NLL $=13.51$, with mutual information $\approx 0$, conditional variance $=0$, and top-1 flip rate $=0$. The internal EVE diagnostics remain interpretable, with layer weights $0.6184/0.1911/0.1905$, per-layer KL $0.2145/0.0587/0.0258$, and latent energies $\mu^2 = 0.1425/0.0163/0.0235$. This compact matched result reinforces the same pattern that appears in the larger comparison below.

\begin{table*}[t]
\centering
\caption{Lower-budget matched deterministic-versus-variational comparison on the dedicated 2.0k-example setting. The same overall pattern already appears at this scale: EVE is stronger on final-validation task metrics and on extended-validation probabilistic metrics.}
\small
\setlength{\tabcolsep}{4pt}
\begin{tabular}{lcccccccc}
\toprule
& \multicolumn{3}{c}{Final validation} & \multicolumn{5}{c}{Extended validation} \\
\cmidrule(lr){2-4} \cmidrule(lr){5-9}
Model & CE$\downarrow$ & PPL$\downarrow$ & Acc$\uparrow$ & MC NLL$\downarrow$ & ECE$\downarrow$ & MI & Flip & CVaR$\downarrow$ \\
\midrule
DET & 5.5417 & 255.12 & 0.1525 & 5.6673 & 0.0603 & $\approx 0$ & 0 & 13.51 \\
EVE & \textbf{5.2896} & \textbf{198.26} & \textbf{0.1752} & \textbf{5.4709} & \textbf{0.0444} & \textbf{0.1830} & \textbf{0.3288} & \textbf{12.51} \\
\bottomrule
\end{tabular}
\end{table*}

We next report the main matched deterministic-versus-variational comparison on about 19,925 raw examples from \textit{WritingPrompts-Filtered}, with GPT-2 tokenization, \texttt{val\_frac=0.2}, batch size 48, and 5 training epochs.

\begin{figure}[t]
\centering
\includegraphics[width=1\linewidth]{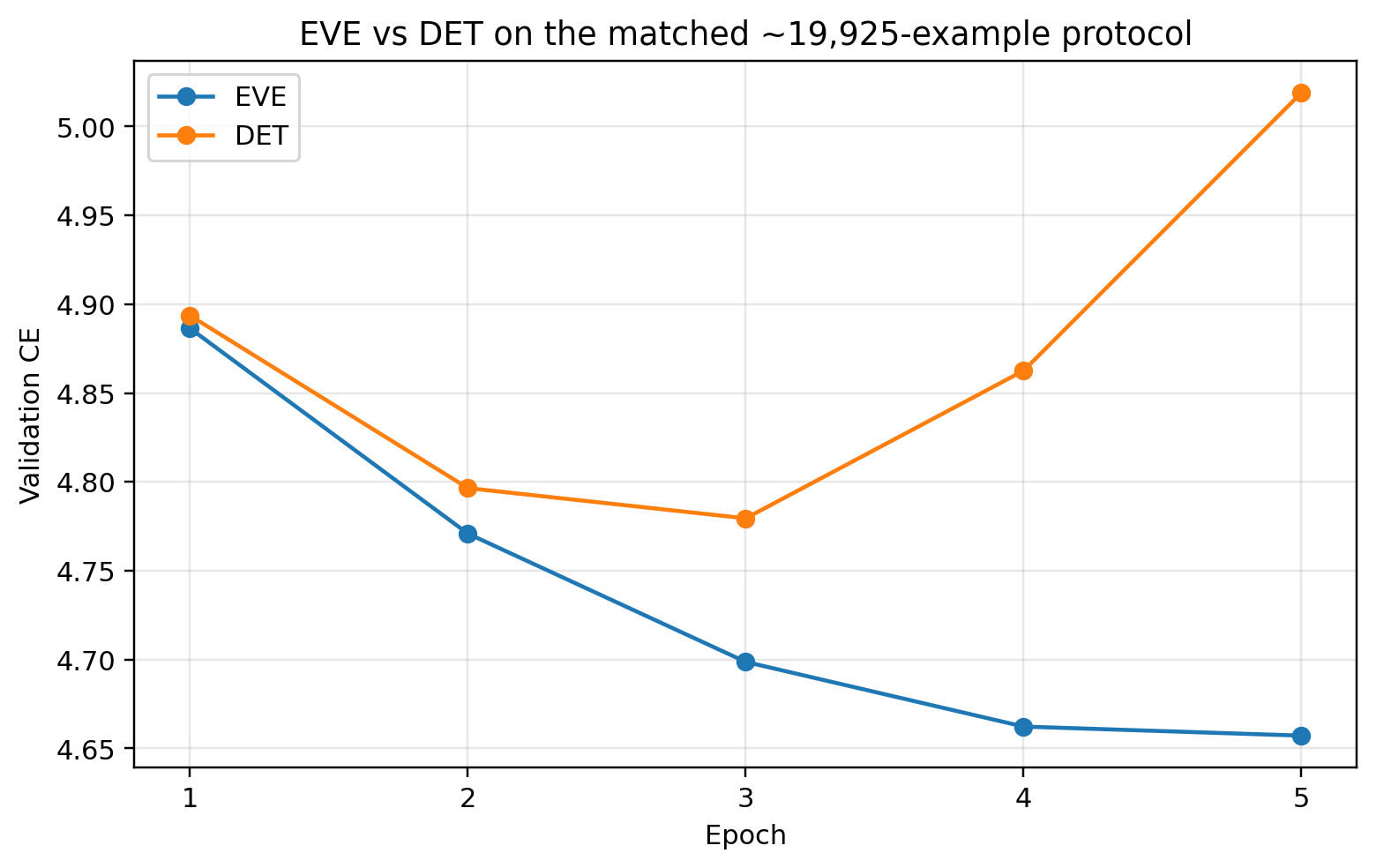}
\caption{Validation CE across epochs on the matched approximately 19,925-raw-example protocol. EVE improves steadily across the full 5-epoch run, whereas DET reaches its best point at epoch 3 and then rises.}
\label{fig:eve-det-ce-curve}
\end{figure}

In \textbf{final validation}, EVE is stronger on all task metrics. EVE selects epoch 5 and reaches CE \(=4.6572\), perplexity \(=105.34\), and accuracy \(=0.2402\), whereas DET selects epoch 3 with CE \(=4.7795\), perplexity \(=119.04\), and accuracy \(=0.2264\). This corresponds to gains of \(0.1223\) in CE, \(13.70\) perplexity points, and about \(1.38\) accuracy points in favor of EVE. The learning dynamics are also cleaner for EVE. Its final-validation CE improves steadily from \(4.8864\) to \(4.6572\) across the full 5-epoch run, while DET reaches its best point at epoch 3 and then rises to \(4.8626\) at epoch 4 and \(5.0190\) at epoch 5, as shown in Figure~\ref{fig:eve-det-ce-curve}. EVE shows sustained validation improvement, whereas DET peaks early and then moves away from its best point.

The \textbf{extended-validation} comparison is also favorable to EVE. EVE reaches Monte Carlo predictive NLL \(=4.6749\) versus \(4.8116\) for DET, and it exhibits a genuine epistemic signal, with mutual information \(=0.1362\), top-1 MC flip rate \(=0.2454\), and epistemic ratio \(=2.81\%\), whereas DET remains near zero on these axes. Calibration is more nuanced. DET achieves the lower ECE, \(0.03546\) versus \(0.05110\), while EVE achieves the lower CVaR-NLL, \(11.8202\) versus \(12.1441\). This gives EVE the stronger tail-risk profile in this matched setting.

\begin{figure}[t]
\centering
\includegraphics[width=1\linewidth]{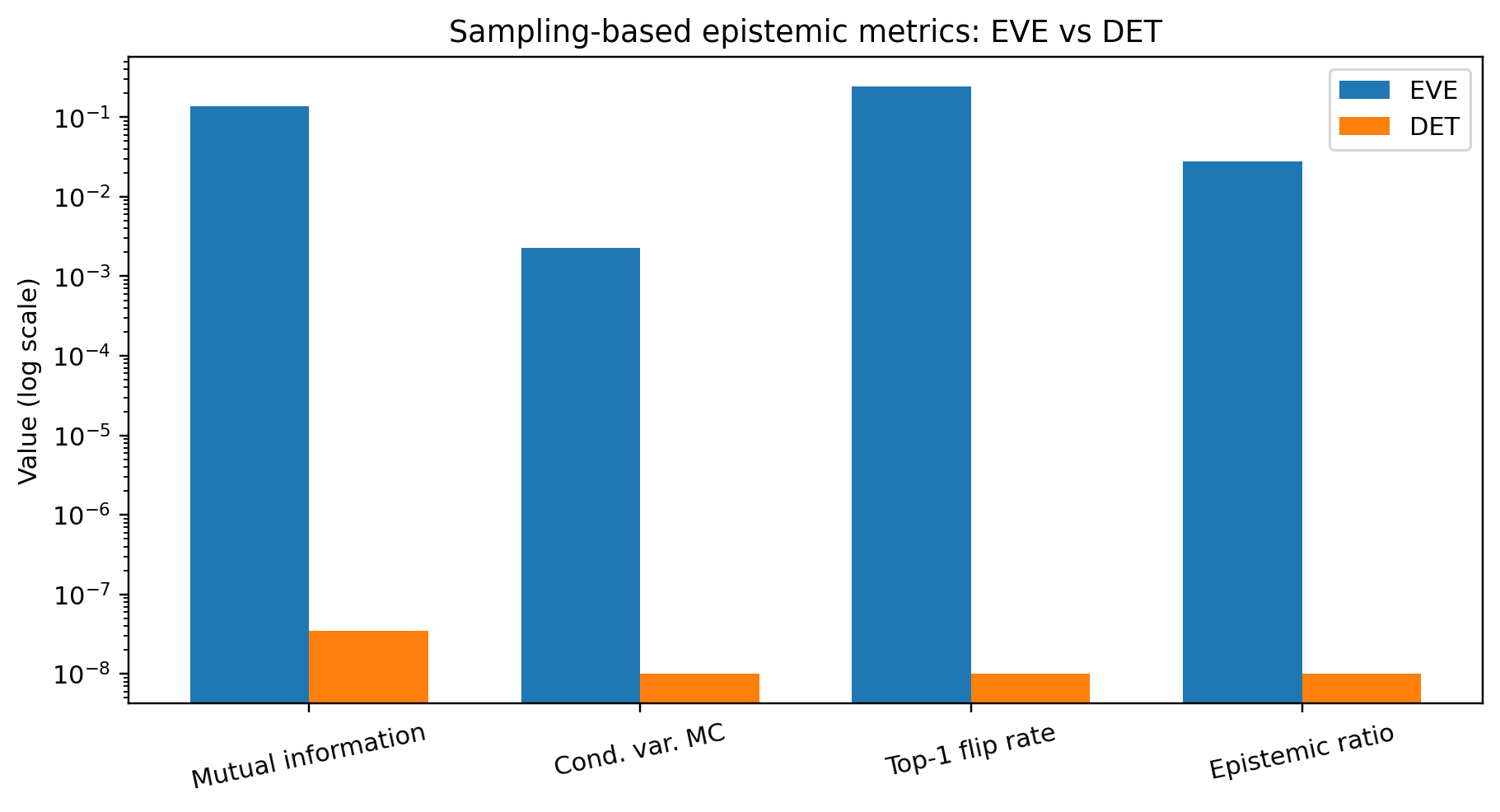}
\caption{Sampling-based epistemic metrics for EVE and the matched deterministic baseline. Mutual information, conditional Monte Carlo variance, top-1 Monte Carlo flip rate, and epistemic ratio are clearly non-zero for EVE, whereas they remain at zero or near-zero values for DET under repeated deterministic forward evaluation. The vertical axis is logarithmic; exact zeros for DET are displayed with a tiny visualization floor only for plotting.}
\label{fig:eve-det-epistemic}
\end{figure}

Figure~\ref{fig:eve-det-epistemic} makes the probabilistic gap explicit. EVE produces non-zero sampling-based epistemic signals, whereas the deterministic baseline remains degenerate on these quantities under repeated deterministic forward evaluation.

From a modeling perspective, this matched comparison shows both stronger predictive quality and richer uncertainty estimates for EVE. In this setting, latent usage remains primarily concentrated in the first layer, with final layer weights \(0.9351/0.0329/0.0320\), layer-1 KL \(=0.18846\), and layer-2 and layer-3 KL values near \(9.4\times10^{-6}\). Both runs are numerically clean, with \texttt{finite\_ok=true} and zero skipped batches, and EVE retains an active homeostatic controller whose final band occupancy remains selective.

Overall, on this matched \(\sim\)19,925-example protocol, EVE is stronger on final-validation CE, perplexity, and accuracy, stronger on extended-validation Monte Carlo predictive NLL and CVaR-NLL, and it provides non-zero epistemic uncertainty. DET retains the lower ECE. EVE therefore combines the stronger task profile with the richer uncertainty-aware profile in this setting.

\section{Discussion}

Our results support a clear conclusion. Local variational neurons integrate well into Transformer feed-forward computation. They preserve strong next-token language-modeling performance and add informative uncertainty signals, including mutual information, conditional variance, and Monte Carlo flip rate.

The experiments separate three useful properties: task quality, useful depth, and internal stability. These properties are related, but they are not the same and should be evaluated separately.

In the 19.9k-example setting, Run B is slightly stronger than Run A on CE, perplexity, and accuracy. Run A, however, retains the richer second-layer latent regime. This shows that stronger task metrics do not automatically mean stronger useful depth. It also shows that latent activity and head usage remain distinct. A latent layer can stay genuinely active even when the final head is dominated by shallower computation.

In the 10.0k-example setting, v23 gives the strongest raw language-modeling point, while v22 gives the cleaner variational regime at essentially no meaningful task-level cost. Relative to v23, v22 reduces $\mu^2_{\text{eval}}$ from $1.5313$ to $0.2119$, $\mu^2_{\text{std,eval}}$ from $53.45$ to $2.94$, and third-layer latent energy from $550.82$ to $13.78$, while keeping CE at $4.8208$, perplexity at $124.06$, and accuracy at $0.2264$. This is a strong practical result. The variational regime can be made much cleaner while remaining competitive on predictive metrics.

The matched deterministic comparisons reinforce the same point. On the dedicated 2.0k-example setting, EVE improves CE from $5.5417$ to $5.2896$, perplexity from $255.12$ to $198.26$, and accuracy from $0.1525$ to $0.1752$, while also producing non-zero epistemic structure under Monte Carlo evaluation. On the matched $\sim$19,925-example protocol, EVE improves CE from $4.7795$ to $4.6572$, perplexity from $119.04$ to $105.34$, accuracy from $0.2264$ to $0.2402$, and Monte Carlo predictive NLL from $4.8116$ to $4.6749$. These results show that variational Transformers can improve task quality while enriching probabilistic behavior.

Overall, variational Transformers already work well at controlled budget. They combine strong language-modeling behavior with explicit internal uncertainty and measurable latent dynamics. This gives a stronger basis for probabilistic evaluation and a practical path toward uncertainty-aware language modeling.

\section{Conclusion}

We introduced a Transformer language model in which deterministic feed-forward units are replaced by local variational neurons. The Transformer backbone remains unchanged, while the internal computation becomes explicitly uncertainty-aware.

Across compact next-token language-modeling experiments, the model remains trainable, numerically stable, and probabilistically informative. It preserves strong task performance, produces informative Monte Carlo uncertainty metrics, and exposes internal latent diagnostics that make depth usage and stability directly measurable.

The main empirical result is direct. Useful uncertainty in language models can be built into the computation itself. Variational Transformers therefore provide a practical path beyond point prediction, toward language models with explicit internal uncertainty structure and stronger probabilistic analysis.

\bibliography{colm2026_conference}
\bibliographystyle{colm2026_conference}

\end{document}